%% file: main.tex
\documentclass{article}
\usepackage{iclr2026_conference,times}

\input{math_commands.tex}

\input{al_header}

\input{color_header}

\newcommand{\ours}{\textsc{GASP}\xspace}
\newcommand{\base}{\textsc{Qwen2.5-Coder-7B}\xspace}

\usepackage{hyperref}
\hypersetup{
    colorlinks,
    linkcolor=newpurple,
    citecolor=newpurple,
    urlcolor=newpurple
}

\usepackage{algorithm}
\usepackage[noend]{algpseudocode}
\usepackage{graphicx}
\usepackage{subcaption}
\usepackage{url}
\usepackage{wrapfig}
\usepackage{tikz}
\usetikzlibrary{arrows.meta,positioning,shapes.geometric}
\usepackage{calligra}

\usepackage{xcolor}
\usepackage{array}
\usepackage{colortbl}

\usepackage{soul}
\definecolor{gaspblue}{HTML}{2F6FB3}
\definecolor{realred}{HTML}{9B363E}
\definecolor{targetbg}{gray}{0.92}
\setul{3pt}{0.8pt}
\newcommand{\bestnoreal}[1]{{\setulcolor{gaspblue}\ul{#1}}}
\newcommand{\bestreal}[1]{{\setulcolor{realred}\ul{#1}}}

\usepackage{multirow}
\usepackage{bigdelim}
\newcommand{\hdr}{\bfseries\footnotesize}

\definecolor{hardred}{HTML}{C62828}
\definecolor{liftorange}{HTML}{f27221}
\definecolor{lemmagreen}{HTML}{307e1a}

\colorlet{hardfill}{hardred!10}
\colorlet{lemmafill}{lemmagreen!10}
\colorlet{liftfill}{liftorange!10}

\title{GASP: Guided Asymmetric Self-Play \\For Coding LLMs}

\author{%
Swadesh Jana\thanks{Equal contribution. Correspondence to: \texttt{swadeshjana@gmail.com} and \texttt{cansu.sancaktar@tue.mpg.de}.
}\, \textsuperscript{$,1$} \; Cansu Sancaktar\footnotemark[1]\,\,\textsuperscript{$,1,2$} \; Tomáš Daniš\textsuperscript{$1$} \; Georg Martius\textsuperscript{$1,2$} \; \\\bfseries  Antonio Orvieto\textsuperscript{$2,3,4$} \; Pavel Kolev\textsuperscript{$1$}\vspace{0.3cm} \\
\textsuperscript{$1$}University of Tübingen \\
\textsuperscript{$2$}Max Planck Institute for Intelligent Systems \\
\textsuperscript{$3$}ELLIS Institute Tübingen \\
\textsuperscript{$4$}Tübingen AI Center \\
\\[2ex]
}

\iclrfinalcopy
\begin{document}

\maketitle

\begin{abstract}
Asymmetric self-play has emerged as a promising paradigm for post-training large language models, where a teacher continually generates questions for a student to solve at the edge of the student's learnability. Although these methods promise open-ended data generation bootstrapped from no human data, they suffer from one major problem: not all problems that are hard to solve are interesting or informative to improve the overall capabilities of the model. Current asymmetric self-play methods are goal-agnostic with no real grounding. We propose \textbf{G}uided \textbf{A}symmetric \textbf{S}elf-\textbf{P}lay (GASP), where grounding is provided by real-data goalpost questions that are identified to pose a hard exploration challenge to the model. During self-play, the teacher first generates an easier variant of a hard question, and then a harder variant of that easier question, with the goal of gradually closing the gap to the goalpost throughout training. Doing so, we improve pass@20 on LiveCodeBench (LCB) by 2.5\% over unguided asymmetric self-play, and through the curriculum constructed by the teacher, we manage to solve hard goalpost questions that remain out of reach for all baselines.
\end{abstract}

\begin{figure*}[!h]
    \centering
    \includegraphics[width=0.96\linewidth]{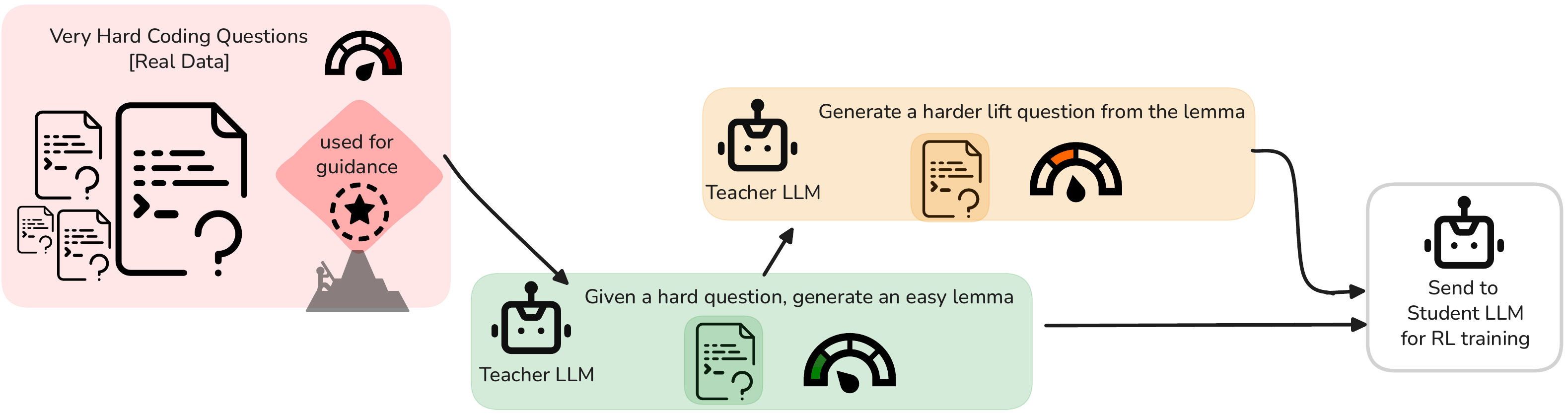}
    \caption{\textbf{Overview of \ours.} Self-play is guided by hard real-data coding questions that a standard RLVR run fails to solve. We refer to this subset as our \textbf{goalpost} questions. The teacher first generates an easy variant (\textbf{\textcolor{lemmagreen}{lemma}}) of this goalpost and then a harder variant (\textbf{\textcolor{liftorange}{lift}}), producing progressively more challenging questions that push the model's knowledge boundary.}
    \label{fig:overview}
\end{figure*}

\section{Introduction}

Asymmetric self-play aims to keep pushing the student model's abilities by generating problems at the frontier of learning. While such methods have recently shown promise for large language model (LLM) post-training \citep{zhao2025absolute}, they quickly run into a fundamental bottleneck: not all hard problems at the agent’s frontier are interesting or help unlock experiences that improve generalization on the downstream tasks we actually care about.

We argue that the teacher needs explicit guidance, i.e. some distant goalpost, to better ground the \textit{interestingness} of the problems it generates. In the prevailing reinforcement learning (RL) post-training paradigm, problems are typically sampled uniformly at random or from a fixed curriculum defined over a static dataset in a given environment \citep{team2025kimi}. In practice, a non-trivial portion of this dataset remains consistently unsolved because it poses a hard exploration challenge. We hypothesize that these genuinely hard, real-data questions, which are known to be difficult yet relevant, form ideal goalposts for \textit{guided} self-play.

In this work, we set out to investigate:
\begin{enumerate}
\item Can we design a guided self-play algorithm that makes meaningful progress on a designated set of hard questions during RL training?
In other words, does steering self-play toward these very hard problems as goalposts actually work?
\item Does such guidance lead the teacher to propose better (i.e., more relevant and informative) problems overall, thereby improving performance on downstream tasks?
\end{enumerate}

We propose \textbf{G}uided \textbf{A}symmetric \textbf{S}elf-\textbf{P}lay (\textbf{GASP}), where the teacher is guided by a set of hard real-data questions $\mathcal{H}$, namely our goalposts.
Given such a target $h \in \mathcal{H}$, the teacher is prompted to generate an easier instance $\ell_0$, which we refer to as the \textbf{lemma}, that aims to preserve the high-level motif of $h$.
We accept a lemma only if it is \emph{learnable but non-trivial} for the student, ensuring that it still provides a clear learning signal unlike the goalpost.
We then prompt the teacher to generate a harder instance $\ell_1$ using $\ell_0$ as a scaffold; we refer to $\ell_1$ as the \textbf{lift} question.
In this way, $\ell_0$ and $\ell_1$ serve as stepping stones that build a curriculum.
Our ultimate goal is to generate questions at the student's learnability frontier that remain relevant and useful for improving the model's coding capabilities, measured through performance on well-known benchmarks.

We evaluate \ours\ on \texttt{LiveCodeBench} (LCB) and show that goalpost guidance improves downstream coding performance. In particular, \ours\ consistently outperforms unguided asymmetric self-play (\textsc{AZR}) and is competitive with standard reinforcement learning with verifiable rewards (RLVR) trained on real-data, with improvements most pronounced at larger $k$.
Beyond benchmark gains, \ours makes progress on the goalpost set itself: as training proceeds, some goalpost questions in $\mathcal{H}$ that remain unsolved by all baselines become solvable.

\section{Related Work}
\paragraph{Asymmetric self-play.}
Asymmetric self-play has been explored in games \citep{sukhbaatar2017_asymmetric} and robotics \citep{openai2021_robotic_selfplay}. More recently, this paradigm has also been applied to RL post-training with LLMs. \citet{poesia2024_minimo} and \citet{dong2025_stp} use teacher--student training for theorem proving. \citet{ye2024_eva} applies similar ideas for alignment, and \citet{liu2025_spice} and \citet{zhang2025better} apply self-play to math and general reasoning. Most notably and closest to our setting, Absolute Zero (\textsc{AZR}) employs asymmetric self-play in the coding domain. \citet{kuba2025language} also tackles asymmetric self-play in instruction-following, math and coding domains, and \citet{teodorescu2023_codeplay} is among early works proposing self-play for coding puzzles, though they only train the teacher.
Recent work by \citet{yu2025guided} also proposes guided self-play for mathematical and general reasoning. However, their grounding relies on few-shot anchoring to a small labeled dataset, rather than steering toward a designated hard set of real-data goalposts as in \ours.
Finally, \citet{wei2025_swerl} proposes self-play for software agents, where an LLM is trained to inject and repair software bugs of increasing complexity in real-world codebases, providing an alternative form of grounding.

\paragraph{Unsupervised environment design and automated curricula.}
A related line of work studies how to automatically generate and schedule training tasks given a parameterized task space.
\citet{parker2022evolving} propose \textsc{ACCEL}, which mutates previously generated levels of the environment and prioritizes levels with high student regret, \ie a gap between the current agent and an optimal agent, to push learning in grid-like game environments. \citet{rutherford2024no} show that commonly used regret approximations can correlate more with success rate than true regret, and propose alternative signals based on student learnability. This perspective is closely related to the teacher objective used in \ours, which encourages generating questions that are challenging yet solvable.

\paragraph{Meta-learning.}
\citet{sundaram2026_soar} propose \textsc{SOAR}, a meta-learning approach that targets hard exploration in RLVR by grounding teacher training in a hard real-data subset in the math domain. At each iteration, the teacher generates synthetic questions for the student, and is rewarded based on the student's measured improvement on the hard subset (used only through this improvement signal). In contrast, \ours\ does not reward the teacher directly for improving on goalpost questions; solving goalposts is instead a byproduct of learning from lemma--lift stepping stones. A practical limitation of the meta-learning objective is that the improvement-based reward can be sparse or noisy when target questions are extremely hard, and it requires repeated inner-loop training and evaluation to obtain the teacher signal.

\section{Method}

We exclusively work in the coding domain, and focus on \texttt{LiveCodeBench} (LCB) \citep{jain2024livecodebench}. We aggregate queries between 2024.10 and 2025.02 as our evaluation set (216 questions; referred to as LCB\textsuperscript{v5}), following \citet{yang2025qwen3}. We treat all queries before 2024.08 as potential goalpost questions and as our training split, which consists of 601 questions. As goalpost questions are filtered exclusively from the training split (pre-2024.08), we ensure there is no contamination with the evaluation set (2024.10 -- 2025.02).

In all our RL experiments and shown baselines, we start from the same base model \base.

\subsection{Background: Standard RLVR vs. Asymmetric Self-Play}
In reinforcement learning with verifiable rewards (RLVR) in the coding domain, we start from a static dataset $\mathcal{D}$ of programming problems. In our case, each problem consists of a natural language description and a set of public example tests, with additional hidden test cases used for evaluation.
At each iteration of RL training, we sample a problem $d \sim \mathcal{D}$ uniformly at random. We then sample a solution from the model $\pi_{\theta}(\cdot \mid d)$, verify it by running all public and private test cases, and assign a binary reward based on pass/fail.

Asymmetric self-play, in contrast, generates problems on-the-go. A single model takes on two roles: (1) the teacher proposing questions and (2) the student solving these questions.
Unless stated otherwise, $\pi_\theta^T$ and $\pi_\theta^S$ share parameters $\theta$ and differ only by role prompting; all updates are applied to the same weights.
The teacher is rewarded based on the student's performance on its proposals: questions should be challenging yet solvable for the student. The student receives rewards based on pass/fail on the questions proposed by the teacher. Absolute Zero (\textsc{AZR}) \citep{zhao2025absolute} follows this paradigm in the coding domain.

In \textsc{AZR}, the teacher generates three types of coding tasks for the student to solve:
\begin{align}
\textbf{Induction:}\quad & \{(i_p,o_p)\}_{p=1}^{P} \;\Rightarrow\; \hat f
&&\text{(infer $f$ from multiple input-output pairs)} \label{eq:f}\\
\textbf{Deduction:}\quad & (f, i) \;\Rightarrow\; \hat o = f(i)
&&\text{(predict output given program and input)} \label{eq:o} \\
\textbf{Abduction:}\quad & (f, o) \;\Rightarrow\; f(\hat i) = o
&&\text{(infer an input consistent with the output)}.
\label{eq:i}
\end{align}
Here, the induction task is the most aligned with typical code contest datasets, where a subset of the input-output pairs proposed by the teacher are used as private test cases (\ie are not shown to the student).

\textsc{AZR} does not start from a human-curated real-world dataset. It makes use of a small dataset of seed questions (256), which are simple generic problems used for bootstrapping.
Our method builds upon \textsc{AZR}, but instead of bootstrapping with seed questions and optimizing the teacher solely based on the student's learnability, \ours provides guidance with \textbf{goalpost} questions.

\subsection{Identifying Candidate Goalpost Problems} \label{sec:goalpost}

Goalpost questions are intended to be hard questions that lie beyond the model's current knowledge boundary and remain unsolved through standard RL training. Our aim is to use these questions to help ground self-play in problems that are relevant for downstream improvement and guide teacher generations.

We identify these goalpost questions via a multi-stage filtering pipeline on the training split (LCB 2023.05 -- 2024.08).
Concretely, we keep only questions that satisfy pass@100$=0$ under all of the following evaluations:
\begin{enumerate}
    \item \textbf{Post-RL filter.} On the base model \base, we run three seeds of RL training and evaluate all checkpoints (saved every 50 iterations) from each run. We keep a question only if pass@100$=0$ for every seed and every checkpoint. We also sample from the base model itself (equivalent to iteration 0 of RL) and discard any question with non-zero pass@100.
    \item \textbf{\textsc{AZR}-checkpoint filter.} We further filter the remaining set using an \textsc{AZR}-trained checkpoint (initialized from \base), again retaining only questions with pass@100$=0$.
    \item \textbf{Final RL filter.} We perform an additional RL run on this hard set to remove any remaining solvable questions.
\end{enumerate}

The final hard set $\mathcal{H}$ consists of 146 problems out of the whole training set (almost 25\%), which we use as goalpost questions for guidance in \ours.

\subsection{\textbf{\ours}: Guided Asymmetric Self-Play}

\begin{wrapfigure}[14]{r}{0.4\textwidth}%
  \vspace{-2.9\baselineskip}
  \centering
\begin{tikzpicture}[x=1cm,y=1cm, font=\sffamily\footnotesize, scale=.57, transform shape]
  \tikzset{
    axis2/.style={{Stealth[length=2.2mm,width=1.6mm]}-{Stealth[length=2.2mm,width=1.6mm]}, line width=0.45pt},
    axis/.style={-{Stealth[length=2.2mm,width=1.6mm]}, line width=0.45pt},
    axis3/.style={-, line width=0.45pt},
    myarrow/.style={-{Stealth[length=2.2mm,width=1.6mm]}, line width=0.5pt},
    dottedarrow/.style={dotted, -{Stealth[length=2.2mm,width=1.6mm]}, line width=0.45pt},
    boundary/.style={dashed, line width=0.45pt},
    shadeRegion/.style={fill=black!10, draw=none},
  }

\path[shadeRegion] (1.2,1.5) rectangle (10.7,7.0);
  \draw[axis]  (1.2,1.2) -- (1.2,7.0);
  \draw[axis3] (0.9,1.5) -- (10.7,1.5);

  \node[rotate=90] at (0.72,4.1) {Difficulty / Complexity};

  \draw[boundary] (1.2,3.62) -- (4.96,3.62);
  \draw[boundary] (5.6,3.62) -- (10.7,3.62);

  \node[anchor=west] at (2.0,6.) {Hard Problem};
  \node[anchor=west] at (6.9,6.55) {\textbf{\textit{Knowledge Space}}};

  \node[anchor=west] at (7.35,4.95) {Expansion};
  \node[anchor=west] at (7.35,4.55) {through Training};

  \node[anchor=west] at (6.85,3.3) {Knowledge Boundary};

  \node[anchor=west] at (4.9,3.74) {Lift};
  \node[anchor=west] at (3.5,2.35) {Lemma};

  \node[star, star points=5, star point ratio=2.25,
        minimum size=10mm, inner sep=0pt, fill=black, draw=black]
        (hard) at (3.2,5.1) {};

  \draw[myarrow] (hard) -- (3.75,2.55);
  \draw[myarrow] (4.65,2.65) -- (5.15,3.45);
  \draw[dottedarrow] (7.2,3.65) -- (7.2,4.65);

\end{tikzpicture}

\caption{\textbf{Illustration of \ours.} Iterative training on generated lemma and lift questions expands the student's knowledge boundary, while the generated questions move closer to the goalpost $h$.}

  \label{fig:knowledge-space}
  \vspace{-0.6\baselineskip}
\end{wrapfigure}
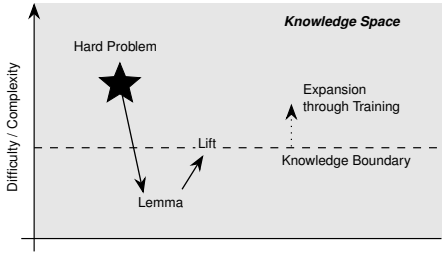

\ours proposes an asymmetric self-play framework that is guided by goalpost questions, where we set out to remedy the goal-agnostic nature of \textsc{AZR}-like approaches and ground the teacher in a known notion of interestingness.

At a given iteration of self-play with \ours, the teacher is shown a hard goalpost question $h_i \in \mathcal{H}$ and is first prompted to generate an easier variant, which we call the \textbf{lemma} question $\ell_0$.
We accept $\ell_0$ only if its estimated pass rate over $N$ trials falls in a learnable band ($0.3\le p\le 0.7$).

Conditioned only on an accepted lemma $\ell_0$, we then prompt the teacher to generate a harder variant of the lemma, which we refer to as the \textbf{lift} question $\ell_1$. Importantly, during lift generation the teacher is not shown the original hard question $h_i$ that serves as the goalpost. This is a deliberate choice: conditioning the lift only on the lemma encourages the teacher to increase difficulty incrementally from the student's current frontier, rather than simply copying surface-level features of the goalpost.

Unlike in \textsc{AZR}, the \ours teacher only generates induction questions (\eqn{eq:f}); abduction and deduction variants are only introduced in the solver phase (\sec{sec:student}). For RL updates, we use Task-Relative REINFORCE++ as proposed in Absolute Zero (\textsc{AZR}) \citep{zhao2025absolute}.

The overall training loop follows three phases (lemma, lift, solver) and is summarized in \alg{alg:gasp-training}.

With more and more iterations of \ours, we repeatedly generate lemma-lift pairs $(\ell_0,\ell_1)$ and train the student on them. This cycle expands the student's knowledge boundary, pushing it upward as shown in \fig{fig:knowledge-space}.
As the student improves, the teacher can generate increasingly difficult lemma and lift questions that are closer to the original goalpost $h$, such that some of these goalpost questions initially out of reach become solvable.

\begin{algorithm}[!h]
\caption{One global iteration of GASP training}
\label{alg:gasp-training}
\begin{algorithmic}[1]
\Require Hard set $\mathcal{H}$; teacher/proposer $\pi_{\theta}^T$ (\ie $\pi_{\theta}$ in teacher role); student/solver $\pi_{\theta}^S$ (\ie $\pi_{\theta}$ in student role); trials $N$;
valid-count target $M$; Global lemma and lift buffers $G_{\ell_0}$, $G_{\ell_1}$ (empty at the start of GASP training)
\vspace{2pt}
\Statex \hrulefill
\State Initialize lemma buffer $\mathcal{B}_0 \gets \emptyset$ \Comment{\textbf{Phase 1: Lemma generation}}
\While{$|\mathcal{B}_0| < M$}
    \State Sample a goalpost $h \sim \mathcal{H}$
    \State $\ell_0 \sim \pi_{\theta}^T(\cdot \mid h)$ \Comment{generate lemma from goalpost}
    \State Estimate $p \gets \textsc{PassRate}(\pi_{\theta}^S, \ell_0, N)$
    \State Compute lemma reward $r_{\text{lemma}}$
    \If{\textsc{IsValid}($\ell_0$) \textbf{and} $0.3 \le p \le 0.7$}
        \State $\mathcal{B}_0 \gets \mathcal{B}_0 \cup \ell_0$
    \EndIf
\EndWhile
\State \textsc{RlUpdate}: Send lemma batch $\mathcal{B}_0$ to trainer and update $\pi_{\theta}^T$
\State $G_{\ell_0} \gets G_{\ell_0} \cup \mathcal{B}_0$                 \Comment{Add local lemma buffer to global buffer for future similarity checks}

\Statex \hrulefill
\vspace{2pt}
\State Initialize lift buffer $\mathcal{B}_1 \gets \emptyset$     \Comment{\textbf{Phase 2: Lift generation}}
\While{$|\mathcal{B}_1| < M$}
    \State Select $\ell_0 \sim \mathcal{B}_0$
    \State $\ell_1 \sim \pi_{\theta}^T(\cdot \mid \ell_0)$ \Comment{generate lift conditioned only on lemma}
    \State Estimate $p \gets \textsc{PassRate}(\pi_{\theta}^S, \ell_1, N)$
    \State Compute lift reward $r_{\text{lift}}$
    \If{\textsc{IsValid}($\ell_1$) \textbf{and} $0.1 \le p \le 0.5$}
        \State $\mathcal{B}_1 \gets \mathcal{B}_1 \cup \ell_1$
    \EndIf
\EndWhile
\State \textsc{RlUpdate}: Send lift batch $\mathcal{B}_1$ to trainer and update $\pi_{\theta}^T$
\State $G_{\ell_1} \gets G_{\ell_1} \cup \mathcal{B}_1$                 \Comment{Add local lift buffer to global buffer for future similarity checks}

\Statex \hrulefill
\vspace{2pt}

\State Construct solver training set $\mathcal{D}_S \gets \mathcal{B}_0 \cup \mathcal{B}_1$ \Comment{\textbf{Phase 3: Solver phase}}
\State Compute solver rewards on $\mathcal{D}_S$ and update solver parameters
\end{algorithmic}
\end{algorithm}

\subsubsection{Teacher Training} \label{sec:teacher}

\paragraph{\textbf{Rewards} }Our teacher / proposer rewards are based on the learnability metric as proposed in \citep{rutherford2024no}. Let $p$ denote the pass rate, \ie the solve rate, of the student across $N$ attempts. Learnability is defined as $p(1-p)^\alpha$ with $\alpha = 1$.

We use the standard learnability reward with $\alpha=1$ for lemma proposals, which peaks at $p=0.5$ and thus favors questions of intermediate difficulty:
\begin{align}
r_{\text{lemma}} =
\begin{cases}
[4 \, p \, (1-p)]^5, & \text{if } 0.3 \leq p \leq 0.7 \\
-0.5, & \text{otherwise}.
\end{cases}
\label{eq:lemma_reward}
\end{align}

\begin{wrapfigure}[14]{r}{0.39\textwidth}%
  \vspace{-.8\baselineskip}
    \includegraphics[width=1\linewidth]{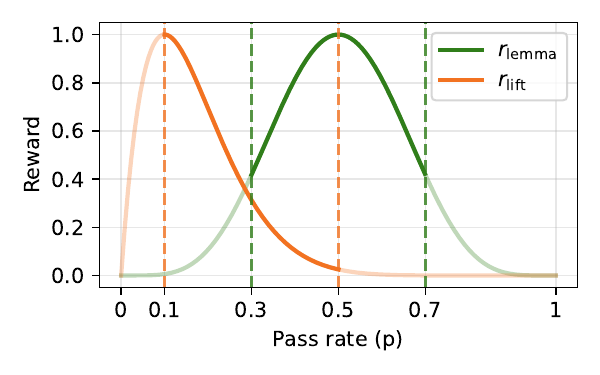}
    \vspace{-.95cm}
  \caption{\textbf{Teacher rewards for \textcolor{lemmagreen}{lemma} and \textcolor{liftorange}{lift} proposals.} We show the learnability curves according to \eqn{eq:lemma_reward} and \eqn{eq:lift_reward}, where the vertical lines mark the corresponding valid bands.}
  \label{fig:rewards}
  \vspace{-1.5\baselineskip}
\end{wrapfigure}

For lift proposals, we use a more skewed learnability reward that peaks at $p=0.1$, favoring harder questions with lower student pass rates:
\begin{align}
r_{\text{lift}} =
\begin{cases}
10 \, p \, \left(\frac{1-p}{0.9}\right)^9, & \text{if } 0.1 \leq p \leq 0.5 \\
-0.5, & \text{otherwise}.
\end{cases}
\label{eq:lift_reward}
\end{align}
\\
The lemma and lift rewards are instances of a generalized learnability reward as further detailed in \app{sec:general_reward}.
The specific exponents and constants are chosen such that the reward curves peak within the desired pass-rate bands and are normalized to [0, 1], as shown in \fig{fig:rewards}.

Additionally, we assign reward $-1$ to any lemma/lift candidate that has a format error.

\paragraph{\textbf{Choosing the difficulty axis}}
In \ours, we parameterize the axis along which we adjust the difficulty of a goalpost question $h$.
An induction-style task is defined by (i) an underlying mapping $f$ to be inferred and (ii) the observed input-output examples used to specify it.

In practice, we distinguish two modes of difficulty adjustment:\\
(1) The \textbf{I/O axis} increases (or decreases) instance or representation complexity while aiming to preserve the same underlying algorithmic motif, \eg by changing the input schema (one list to multiple lists or nested lists) or by selecting examples that make the function $f$ harder to infer.\\
(2) The $\mathbf{f}$ \textbf{axis} increases (or decreases) algorithmic complexity by modifying the mapping itself, \eg by introducing additional constraints or composing new operations that require extra logic beyond the original rule.

For each lemma generation, we uniformly sample which axis to apply (I/O or $f$).
The corresponding lift question then aims to increase the difficulty along the same axis.

\paragraph{\textbf{Rejection sampling for lemma-lift proposals}} Each lemma/lift candidate returned by the teacher undergoes lightweight validity checks before it is passed to the solver and admitted to the training buffer (in addition to the pass-rate band checks in Algorithm~\ref{alg:gasp-training}).
We follow prior work \citep{zhao2025absolute} and reject proposals that are malformed, unsafe to execute, or exhibit non-deterministic behavior under repeated runs.

In addition, we enforce \textbf{novelty} with respect to the \textbf{global buffer} of previously accepted lemma/lift proposals.
Concretely, we compute cosine similarity between embeddings of the proposed question text and generated code and those of previously accepted items.
If the similarity exceeds $0.95$ for any item in either buffer, the proposal is rejected (see \app{sec:buffer_dis} for more details).
This diversity filter prevents mode collapse and encourages the teacher to generate a broad set of distinct lemma/lift questions.

\subsubsection{Student Training} \label{sec:student}
For valid lemma and lift proposals produced in the teacher phase, we train the student using verifiable rewards based on pass/fail. For each training question, we choose uniformly at random whether to keep it in the induction format or to convert it into a deduction or abduction format. In the latter case, instead of presenting multiple input-output examples as in induction, we present the function $f$ together with either a single input and ask the student to predict the corresponding output (deduction, \eqn{eq:o}), or a single output and ask the student to produce a corresponding input (abduction, \eqn{eq:i}). Since \textsc{AZR} reported gains from including deduction and abduction tasks, we adopt the same strategy to maintain diversity in the student training phase.

\subsection{\ours with Real-data RL} \label{sec:joint}

We also test a variant of \ours\ that performs joint training on real-data and synthetic data generated through guided asymmetric self-play. After completing one \ours\ iteration as detailed in Algorithm~\ref{alg:gasp-training}, we additionally sample questions from the real-data training split $\mathcal{D}$, corresponding to the \texttt{LiveCodeBench} time window 2023.05 -- 2024.08, and apply RLVR updates on these samples. Note that this is the same split used to construct our goalpost set, such that $\mathcal{H} \subset \mathcal{D}$.

\section{Results}

We evaluate \ours\ and report pass@k on our \texttt{LiveCodeBench} evaluation split (LCB\textsuperscript{v5}). We also evaluate \ours + Real-data RL, the joint-training variant described in Section~\ref{sec:joint}, and a Real-data RL baseline that applies standard RLVR to the static LCB training split. We further compare against the base model \base\ and \textsc{AZR}, \ie unguided self-play.

We find that \ours yields significant gains over the base model and consistently outperforms \textsc{AZR}, supporting the hypothesis that goalpost guidance improves the quality of self-play training signal.
\ours is broadly competitive with Real-data RL across $k$, with slightly higher mean pass@k at larger $k$, though the difference is modest.
Moreover, \ours + Real-data RL improves over \ours, indicating that guided self-play is effective on its own, while also being complementary to real-data training and able to benefit from additional real-world supervision when available. We also evaluate greedy pass@1 (temperature 0) on LCB\textsuperscript{v5}, HumanEval$^+$~\citep{liu2023your}, and MBPP$^+$~\citep{liu2023your} (see \tab{tab:greedy-results}). On the target distribution (LCB\textsuperscript{v5}), \ours and \ours + Real-data RL maintain their advantage. On HumanEval$^+$, \ours closely follows \textsc{AZR}, while \ours + Real-data RL and \ours lead on MBPP$^+$.

  \begin{figure*}[t]
    \centering
    \includegraphics[width=0.7\linewidth]{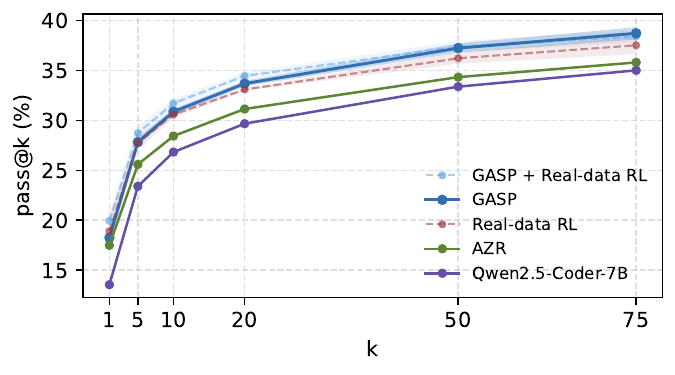}
   \vspace*{-.35cm}
    \caption{\textbf{Pass@k performance on the LCB eval benchmark (LCB\textsuperscript{v5}).} We compare \ours, \ours + Real-data RL, Real-data RL, \textsc{AZR}, and \base. All RL-based results are repeated over three seeds.
    For each seed of our RL runs, we perform single-checkpoint model selection by choosing the checkpoint that maximizes pass@20 on the LCB\textsuperscript{v5} evaluation split, and report mean $\pm$ std across seeds.
    For \textsc{AZR}, we evaluate the authors' publicly released checkpoint (trained from \base); the authors report selecting their checkpoint based on best performance on their evaluation benchmarks (including LCB) over training from a single-seed run. We additionally report pass@1-based checkpoint selection in the appendix (Figure~\ref{fig:gasp_rj_best_by_1}), which shows the same qualitative trends.}
    \label{fig:res_best_seed20}
      \vspace{.2cm}
      \centering
\captionof{table}{\textbf{LCB\textsuperscript{v5} pass@1 and pass@20 results.} We use bold for \textbf{global best}, blue underline for \bestnoreal{best self-play} and red for \bestreal{best with real-data training}.}
\vspace{.1cm}
\label{tab:lcb-main}

\begin{tabular}{lcccc}
\toprule
& \multirow{2}{*}{\shortstack{Real-data\\Training}}
& \multirow{2}{*}{\shortstack{Real-data\\Guidance}}
& \multicolumn{2}{c}{LCB$^{\text{v5}}$} \\
\cmidrule(lr){4-5}
& & & pass@1 & pass@20 \\
\midrule
GASP (ours) & \xmark & \cmark & \bestnoreal{18.26$_{\pm0.68}$} & \bestnoreal{33.69$_{\pm0.28}$} \\
\textsc{AZR} & \xmark & \xmark & 17.49$_{\phantom{\pm0.00}}$ & 31.15$_{\phantom{\pm0.00}}$ \\
\midrule
Real-data RL & \cmark & \xmark & 18.91$_{\pm1.40}$ & 33.10$_{\pm0.12}$ \\
GASP + Real-data RL (ours) & \cmark & \cmark & \bestreal{\textbf{19.93$_{\pm0.88}$}} & \bestreal{\textbf{34.46$_{\pm0.34}$}} \\
\midrule
Qwen2.5-Coder-7B & --- & --- & 13.55$_{\phantom{\pm0.00}}$ & 29.68$_{\phantom{\pm0.00}}$ \\
\bottomrule
\end{tabular}
  \end{figure*}

\paragraph{\textbf{Do we need rejection sampling?}}
As described in \sec{sec:teacher}, we perform a rejection sampling step in \ours to ensure that the generated lemma and lift proposals maintain sufficient diversity. We run an ablation of \ours without rejection sampling and report pass@k eval performance on LCB\textsuperscript{v5} in \Fig{fig:gasp_no_rejection_sampling}. Although \ours without rejection sampling still outperforms \textsc{AZR} and shows comparable performance with real-data RL, we observe high variance across seeds.

Upon inspection of generated lemma and lift proposals, we can attribute this high variance to diversity collapse in some of the runs. By rejecting proposals that are overly similar to previously accepted ones, rejection sampling mitigates this failure mode and yields more consistent gains.

\paragraph{\textbf{Can we solve the goalpost hard questions with \ours?}}
\begin{figure}[t]
    \centering
    \begin{subfigure}[t]{0.49\textwidth}
        \centering
        \includegraphics[width=\linewidth]{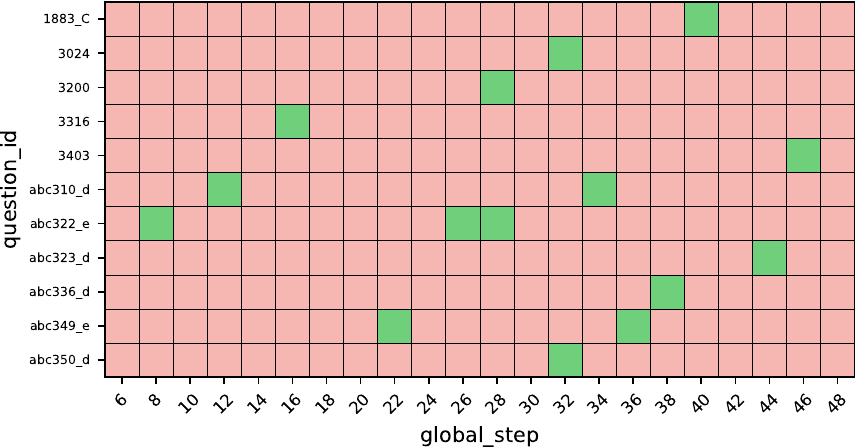}
        \caption{GASP}
        \label{fig:res_best_seed_gasp}
    \end{subfigure}
    \begin{subfigure}[t]{0.49\textwidth}
        \centering
        \includegraphics[width=\linewidth]{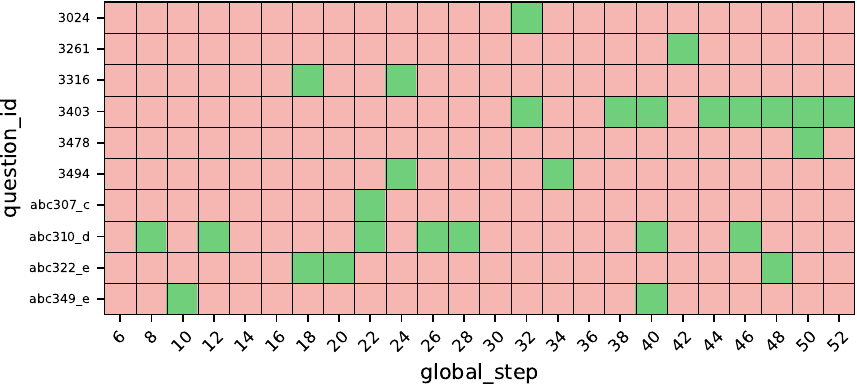}
        \caption{GASP + Real-data RL}
        \label{fig:res_best_seed_real}
    \end{subfigure}\hfill
\caption{\textbf{Goalpost questions solved with \ours and \ours + Real-data RL.} Visualization over training checkpoints (x-axis: global step) and goalpost question IDs (y-axis). We include only goalpost questions that are solved at least once during training. A cell is marked as solved (green) if pass@$100 > 0$ (i.e., at least one of 100 samples passes); otherwise it is unsolved (red). Results are aggregated across three seeds, where a cell is marked solved if \emph{any} seed solves the question at that checkpoint (union across seeds).}
    \label{fig:goalpost_gasp}
\end{figure}

As lemma and lift questions serve as stepping stones toward each goalpost, \ours can provide a curriculum that makes progress on questions in $\mathcal{H}$.
By construction, these goalposts remain unsolved throughout the filtering pipeline in \sec{sec:goalpost}, and a standard RLVR run trained only on $\mathcal{H}$ also fails to solve any goalpost question over the course of training. Notably, the publicly released AZR checkpoint also solves none of the 146 goalpost questions.

Evaluating checkpoints throughout RL training, \ours solves 11 unique goalpost questions out of 146 (6/3/3 per seed, union of 11 across three seeds), with all seeds making non-trivial progress on questions unsolvable by any baseline. In Figure~\ref{fig:goalpost_gasp}, we visualize when each goalpost becomes solved across training, marking a goalpost as solved at a given checkpoint if it is solved by \emph{any} seed. We report the corresponding per-seed counts for the \ours\ variants (without rejection sampling (\fig{fig:gasp_no_rj_goalpost}) and \ours + Real-data RL) in \supp{app:goalpost} and \tab{tab:hard-evals}.

We observe that goalpost questions are solved intermittently rather than persistently across evaluated checkpoints: a question solved at one checkpoint can be unsolved at the next. This is consistent with these questions lying at the frontier of the model's capability, where the correct reasoning path and answer lie in the tail of the sampling distribution. As GASP + Real-data RL showcases more consistent solving across checkpoints, one could argue that additional RL training on real data, which also includes the goalpost questions themselves in the training split, helps consolidate the capabilities unlocked by asymmetric self-play.

We ablate two further design choices: the input-output difficulty axis and the two-stage curriculum in \ours.

\paragraph{Difficulty axis ablation for the teacher.} As described in \sec{sec:teacher}, \ours uniformly samples between two modes of difficulty adjustment: the I/O axis (input--output complexity) and the $f$ axis (algorithmic complexity). Removing the I/O axis in \ours, such that all difficulty adjustments operate only along the $f$ axis, leads to comparable pass@$k$ performance on LCB\textsuperscript{v5}, but with substantially increased variance among different seeds. We also observe worse goalpost performance (4 unique questions solved vs.\ 11 for \ours), suggesting that varying input--output complexity helps inject extra diversity into self-play, which is important for generating an effective curriculum. Full ablation results can be found in \app{sec:io-ablation}.

  \paragraph{Two-stage curriculum ablations.} Replacing the lemma--lift curriculum with a single-stage reward targeting intermediate difficulty (one-step medium, reward peak at student pass rate $p{=}0.3$) leads to worse pass@$k$ on LCB\textsuperscript{v5} compared to \ours, while targeting only hard questions (one-step hard, peak at student pass rate $p{=}0.1$) performs slightly better but still lags behind \ours. On goalpost questions, one-step hard solves only 2 unique questions across 3 runs, suggesting that medium-difficulty stepping stones are important for an effective curriculum. Still, both one-step ablations outperform \textsc{AZR} at higher $k$, highlighting the value of guided synthetic data
  generation even without the full two-stage curriculum. See \app{sec:one-step-ablation} for more details.

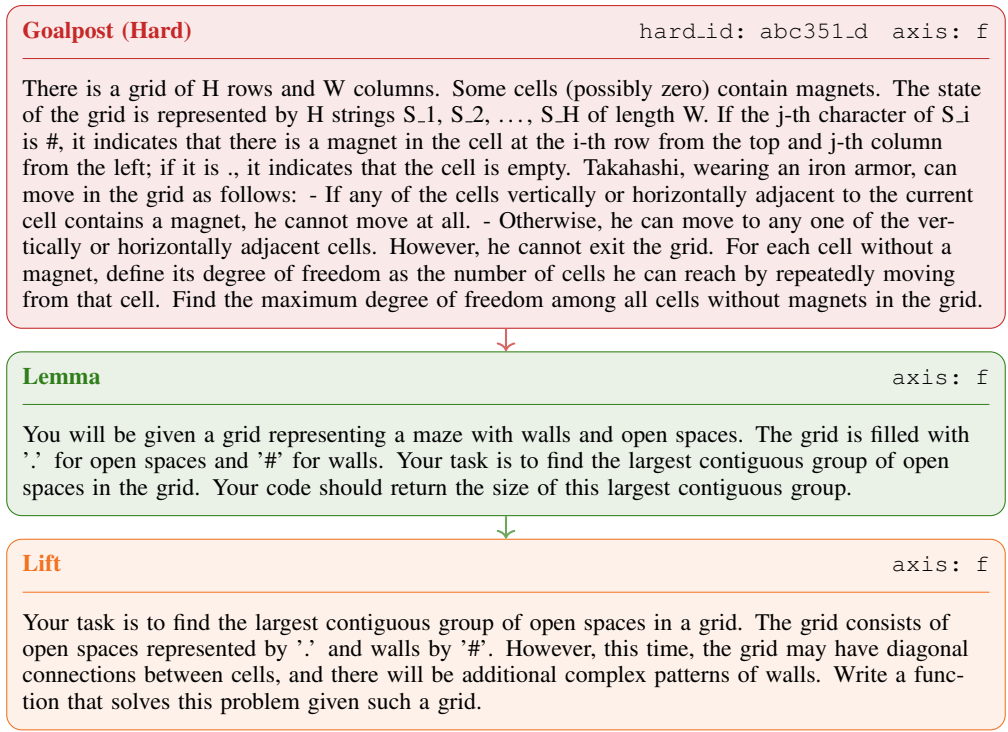
\begin{figure}[t]
\centering

\begin{tikzpicture}[
  font=\small,
  box/.style={draw, rounded corners=2mm, align=left, inner sep=6pt, text width=0.92\linewidth},
  hdr/.style={font=\bfseries\footnotesize}
]

\node[box, draw=hardred, fill=hardfill, anchor=north] (hard1) {
  {\hdr\textcolor{hardred}{Goalpost (Hard)}}\hfill\texttt{hard\_id: abc351\_d}\quad\texttt{axis: f}\\[-2pt]
  \textcolor{hardred}{\rule{\linewidth}{0.3pt}}\\[4pt]
There is a grid of H rows and W columns. Some cells (possibly zero) contain magnets.
The state of the grid is represented by H strings S\_1, S\_2, \ldots, S\_H of length W. If the j-th character of S\_i is \#, it indicates that there is a magnet in the cell at the i-th row from the top and j-th column from the left; if it is ., it indicates that the cell is empty.
Takahashi, wearing an iron armor, can move in the grid as follows:
- If any of the cells vertically or horizontally adjacent to the current cell contains a magnet, he cannot move at all.
- Otherwise, he can move to any one of the vertically or horizontally adjacent cells.
However, he cannot exit the grid.
For each cell without a magnet, define its degree of freedom as the number of cells he can reach by repeatedly moving from that cell. Find the maximum degree of freedom among all cells without magnets in the grid.
};

\node[box, draw=lemmagreen, fill=lemmafill, below=3mm of hard1] (lemma1) {
  {\hdr\textcolor{lemmagreen}{Lemma}}\hfill\texttt{axis: f}\\[-2pt]
  \textcolor{lemmagreen}{\rule{\linewidth}{0.3pt}}\\[4pt]
You will be given a grid representing a maze with walls and open spaces. The grid is filled with '.' for open spaces and '\#' for walls. Your task is to find the largest contiguous group of open spaces in the grid. Your code should return the size of this largest contiguous group.
};

\node[box, draw=liftorange, fill=liftfill, below=3mm of lemma1] (lift1) {
  {\hdr\textcolor{liftorange}{Lift}}\hfill\texttt{axis: f}\\[-2pt]
  \textcolor{liftorange}{\rule{\linewidth}{0.3pt}}\\[4pt]
Your task is to find the largest contiguous group of open spaces in a grid. The grid consists of open spaces represented by '.' and walls by '\#'. However, this time, the grid may have diagonal connections between cells, and there will be additional complex patterns of walls. Write a function that solves this problem given such a grid.
};

\draw[->, thick, hardred!70] (hard1.south) -- (lemma1.north);
\draw[->, thick, lemmagreen!70] (lemma1.south) -- (lift1.north);

\end{tikzpicture}

\caption{\textbf{Goalpost$\rightarrow$lemma$\rightarrow$lift examples from \ours run.} Functional-axis ($f$) selected for difficulty adjustment on hard question with id \texttt{abc351\_d}. (shared motif: connectivity/reachability)}

\label{fig:lemma-lift-example-3382}
\end{figure}

We also provide some qualitative examples of the generated lemma and lift questions as shown in \fig{fig:lemma-lift-example-3382} (more examples, including i/o difficulty axis, shown in \fig{fig:lemma-lift-more} and \fig{fig:lemma-lift-more-f}).

\section{Discussion}

We propose guided self-play with \ours and showcase that goalpost guidance helps generate questions that are not just progressively more difficult but also more relevant for downstream improvement. \ours consistently outperforms goal-agnostic self-play with \textsc{AZR}.

As we take a known hard problem, first making it easier through the \textit{lemma} question and then harder again with the \textit{lift}, we provide a curriculum through these stepping stones. As a result, we are able to solve some of these difficult questions that originally posed a hard-exploration challenge in RL, providing no learning signal on their own.

We find that \ours performs strongly without additional real-data RL updates, while the joint variant \ours + Real-data RL yields further improvements on LCB\textsuperscript{v5}. This indicates that guided self-play can be used for standalone training, but also complements standard RLVR pipelines when real-data supervision is available. The rejection sampling step improves stability by maintaining diversity among lemma/lift proposals, helping prevent overfitting and mode collapse.

\vspace{-.8em}
\paragraph{\textbf{Limitations}}
We provide a proof of concept of guided asymmetric self-play in the coding domain, showing promising results, but there are several clear avenues for improvement. First, beyond the diversity-based similarity checks, we do not explicitly validate that the teacher's proposed lemma/lift questions provide \emph{correct} guidance toward the goalpost. As a result, the generated stepping stones may not always align with the target hard question in an obvious way. Similarly, when selecting the difficulty axis, we observe that the input-output axis is often harder for the teacher, and the teacher can default back to increasing functional difficulty, which we do not penalize. Lightweight checks that measure alignment to the goalpost, or introducing a judge/reward model, could address both issues.\looseness=-1

Second, we observe a more fundamental failure mode: current LLMs are not always good at abstracting the underlying concept of a task. They often borrow surface-level metaphors from the original problem rather than producing clean conceptual stepping stones. Relatedly, when asked to increase difficulty, they frequently do so by adding extra constraints, which is not always the most informative form of complexity. Since models can often \emph{judge} stepping-stone quality better than they can generate it, incorporating judge-based rewards can be a future direction.

Additionally, we use LCB pre-2024.08 as our training split, which is our sole real-data source. Scaling GASP with larger code training corpora is a natural next step.

Finally, our framework uses a fixed set of goalposts. An open question is what should happen once a goalpost is reached: ideally, the goalpost set should be updated over time so that guidance remains meaningful as the model improves. We leave dynamic goalpost updating to future work.

\subsubsection*{Acknowledgments}
The authors gratefully acknowledge the computing resources provided by Max Planck Institute for Intelligent Systems (MPI-IS) and thank the International Max Planck Research School for Intelligent Systems (IMPRS-IS) for supporting Cansu Sancaktar and Tomáš Daniš.
Antonio Orvieto acknowledges financial support from the Hector Foundation.
Georg Martius is a member of the Machine Learning Cluster of Excellence, EXC number 2064/1 – Project number 390727645.
This work was supported by the ERC - 101045454 REAL-RL and the German Federal Ministry of Education and Research (BMBF): Tübingen AI Center, FKZ: 01IS18039A.

\bibliography{refs}
\bibliographystyle{iclr2026_conference}

\newpage
\appendix
\section{Appendix}

\subsection{Generalized Learnability Reward}
\label{sec:general_reward}

The lemma and lift rewards (\eqn{eq:lemma_reward}, \eqn{eq:lift_reward}) are instances of a generalized family of learnability rewards:
\begin{equation}
\label{eq:generalized_reward}
\mathcal{R}(p;\, a, b) = \left[\frac{p}{a}\left(\frac{1-p}{1-a}\right)^{\frac{1-a}{a}}\right]^b,
\qquad p \in [0,1],\; a \in (0,1),\; b \in \mathbb{R}.
\end{equation}
This function attains its maximum at $p = a$, where $a$ controls the target difficulty and $b$ controls the sharpness of the peak. The lemma reward corresponds to $(a{=}0.5, b{=}5)$ and the lift reward to $(a{=}0.1, b{=}1)$.

By construction, the reward is normalized such that its maximum value satisfies $\mathcal{R}^*(p;\cdot) = 1$. This normalization ensures that different reward configurations remain comparable in magnitude while differing only in the region of the difficulty space that they prioritize.

\section{Buffer Dissimilarity}
\label{sec:buffer_dis}
We monitor \emph{buffer dissimilarity}, which measures how newly generated samples differ from previously generated ones stored in a buffer.

Let $\mathcal{B}_t$ denote the batch at iteration $t$ and $\mathcal{M}_{t-1}$ denote the accumulated buffer from earlier iterations. Buffer dissimilarity is defined as
\begin{equation}
D_{\mathrm{buffer}}(\mathcal{B}_t, \mathcal{M}_{t-1})
= \frac{1}{|\mathcal{B}_t|} \sum_{x \in \mathcal{B}_t}
\left( \frac{1}{|\mathcal{M}_{t-1}|} \sum_{x' \in \mathcal{M}_{t-1}} \big(1 - \mathrm{sim}(x, x')\big) \right).
\end{equation}

This metric provides a notion of temporal novelty. High buffer dissimilarity indicates continued expansion into previously unexplored regions of the task space, whereas decreasing values suggest increasing reuse of rewarded patterns.

\section{Extended Results}

\begin{figure*}[!h]
    \centering
    \includegraphics[width=0.7\linewidth]{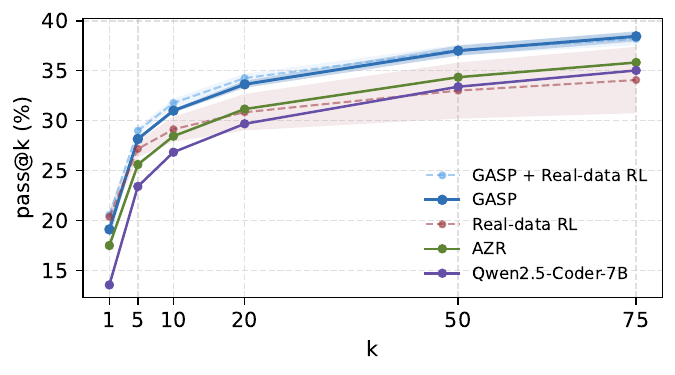}
    \caption{\textbf{Pass@k performance on the LCB eval benchmark (LCB\textsuperscript{v5}) with checkpoint selected for best pass@1.} We compare \ours, \ours + Real-data RL, Real-data RL, \textsc{AZR}, and \base. All RL-based results are repeated over three seeds. For each seed of our RL runs, we select the checkpoint with the best pass@1 within that run, and report mean $\pm$ std across seeds. For \textsc{AZR}, we evaluate the authors' publicly released checkpoint (trained from \base); the authors report selecting the best checkpoint over training from a single-seed run.}

    \label{fig:gasp_rj_best_by_1}
\end{figure*}

We show a detailed view of the evaluation results on LCB\textsuperscript{v5} for pass@1 and pass@20 in \tab{tab:lcb-joint-ablations}. All pass@k evaluations use temperature 0.6 with $N=100$ samples for the pass@k estimator. We also provide pass@k curves for RL runs, where we select each run's checkpoint by maximizing pass@1 in \Fig{fig:gasp_rj_best_by_1}. Under pass@1-based checkpoint selection, \ours exhibits a clearer separation from Real-data RL.

\begin{wrapfigure}[15]{r}{0.41\textwidth}%
  \vspace{-1\baselineskip}
    \includegraphics[width=1\linewidth]{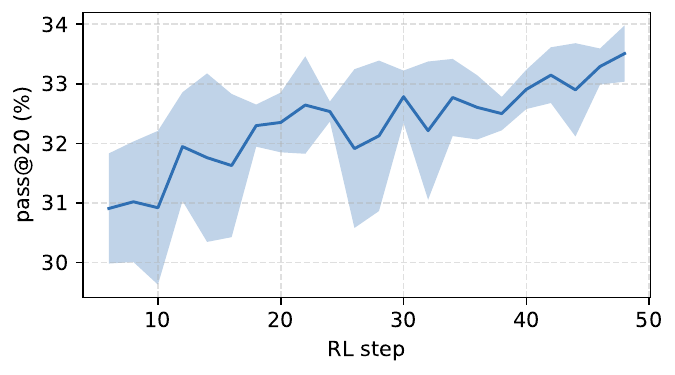}
    \vspace{-.75cm}
  \caption{Pass@20 performance on the \textbf{LCB\textsuperscript{v5} benchmark throughout RL training for \ours.} Evaluation performance continues to improve throughout training and shows no plateau by the final iteration, where the runs were truncated due to compute constraints.}
  \label{fig:training_curve}
  \vspace{-1.5\baselineskip}
\end{wrapfigure}
Note that all Real-data RL runs were trained until convergence. As reported by \citet{zhao2025absolute}, \textsc{AZR} (with \base) was run for 350 global steps, after which the best checkpoint was chosen according to overall performance on the math and code evaluation benchmarks (including \texttt{LiveCodeBench}). \ours with rejection sampling was run for approximately 50 global iterations due to compute budget limitations.
As shown in \fig{fig:training_curve}, performance on LCB\textsuperscript{v5} was still improving at the end of training, suggesting
  further gains are possible with additional compute. Each iteration is more costly than without rejection sampling, as the teacher must generate multiple proposals until a sufficiently novel one is accepted.
These gains are achieved despite substantially fewer training iterations than the baselines.

In \tab{tab:greedy-results}, we report greedy pass@1 (temperature 0) on LCB\textsuperscript{v5}, HumanEval$^+$, and MBPP$^+$. HumanEval$^+$\footnote{HumanEval$^+$ extends HumanEval~\citep{chen2021evaluating}, a benchmark of 164 Python function completion tasks given a signature and docstring, with more comprehensive test cases.} and MBPP$^+$\footnote{MBPP$^+$ extends the Mostly Basic Python Problems (MBPP) benchmark~\citep{austin2021program} with more comprehensive test cases (378 questions).} are substantially more saturated benchmarks. On HumanEval$^+$, the smallest benchmark with 164 problems, \textsc{AZR} leads, though all methods remain within a narrow margin. On MBPP$^+$, all methods perform comparably, confirming that self-play training does not degrade general coding ability.

\begin{table}[h]
\centering
\caption{\textbf{Performance on the LCB\textsuperscript{v5} benchmark.} We compare \ours pass@1 and pass@20, as well as ablations of our method, with standard RLVR training on all LCB questions in our training time split (Real-data RL), as well as performance of the \textsc{AZR} checkpoint open-sourced by \citet{zhao2025absolute}, and the base model \base. We show performance across 3 seeds of RL training respectively for two selection criteria: in each seed run we choose the model checkpoint with the best pass@1 or pass@20 performance. \ours outperforms \textsc{AZR} (unguided self-play) across all selection criteria. We use bold for \textbf{global best}, blue underline for \bestnoreal{best self-play} and red for \bestreal{best with real-data training}.}
\label{tab:lcb-joint-ablations}
\resizebox{\textwidth}{!}{
\begin{tabular}{clcccccc}
\toprule
& & \multirow{2}{*}{\shortstack{Training on\\Real-data}}
& \multirow{2}{*}{\shortstack{Real-data\\Guidance}}
& \multicolumn{2}{c}{\textit{selected by pass@1}}
& \multicolumn{2}{c}{\textit{selected by pass@20}} \\
\cmidrule(lr){5-6} \cmidrule(lr){7-8}
& & & & pass@1 & pass@20 & pass@1 & pass@20 \\
\midrule
 & GASP & \xmark & \cmark & 19.10$_{\pm0.92}$ & \bestnoreal{33.63$_{\pm0.35}$} & 18.26$_{\pm0.68}$ & \bestnoreal{33.69$_{\pm0.28}$} \\
\ldelim\{{4}{30pt}[\scriptsize\textit{Ablations}] & GASP no i/o axis & \xmark & \cmark & \bestnoreal{19.36$_{\pm1.35}$} & 33.31$_{\pm1.34}$ & \bestnoreal{18.94$_{\pm1.69}$} & 33.47$_{\pm1.14}$ \\
 & GASP w/o rejection sampling & \xmark & \cmark & 17.97$_{\pm0.79}$ & 32.72$_{\pm1.81}$ & 17.64$_{\pm1.33}$ & 33.03$_{\pm1.50}$ \\
 & GASP one-step medium & \xmark & \cmark & 17.10$_{\pm0.33}$ & 31.41$_{\pm0.03}$ & 15.73$_{\pm1.61}$ & 32.36$_{\pm0.03}$ \\
 & GASP one-step hard & \xmark & \cmark & 17.76$_{\pm1.14}$ & 32.20$_{\pm0.91}$ & 17.13$_{\pm0.68}$ & 33.37$_{\pm0.12}$ \\
 & \textsc{AZR} & \xmark & \xmark & 17.49$_{\phantom{\pm0.00}}$ & 31.15$_{\phantom{\pm0.00}}$ & 17.49$_{\phantom{\pm0.00}}$ & 31.15$_{\phantom{\pm0.00}}$ \\
\midrule
 & Real-data RL & \cmark & \xmark & 20.35$_{\pm0.75}$ & 30.84$_{\pm1.83}$ & 18.91$_{\pm1.40}$ & 33.10$_{\pm0.12}$ \\
 & GASP + Real-data RL & \cmark & \cmark & \bestreal{\textbf{20.55$_{\pm0.26}$}} & \bestreal{\textbf{34.25$_{\pm0.45}$}} & \bestreal{\textbf{19.93$_{\pm0.88}$}} & \bestreal{\textbf{34.46$_{\pm0.34}$}} \\
\midrule
\scriptsize\textit{Base} & Qwen2.5-Coder-7B & --- & --- & 13.55$_{\phantom{\pm0.00}}$ & 29.68$_{\phantom{\pm0.00}}$ & 13.55$_{\phantom{\pm0.00}}$ & 29.68$_{\phantom{\pm0.00}}$ \\
\bottomrule
\end{tabular}
}
\end{table}

\begin{table}[t]
\centering
\caption{\textbf{Greedy pass@1 performance on LCB\textsuperscript{v5}, HumanEval$^+$ and MBPP$^+$.} Checkpoint selected by best LCB pass@20. Mean over seeds; $\pm$ shows standard deviation.
For \textsc{AZR}, we evaluate the authors' publicly released checkpoint (trained
from \base); the authors report selecting the best checkpoint based on performance on their evaluation benchmarks, including LCB\textsuperscript{v1-5}, HumanEval$^+$, and MBPP$^+$ over training from a single-seed run.
\colorbox{targetbg}{Shaded area} indicates our target distribution, \ie LCB\textsuperscript{v5}. We use bold for \textbf{global best}, blue underline for \bestnoreal{best self-play} and red for \bestreal{best with real-data training}.}
\label{tab:greedy-results}
\begin{tabular}{l>{\columncolor{targetbg}}ccc}
\toprule
Method & LCB$^{\text{v5}}$ & HumanEval$^+$ & MBPP$^+$ \\
\midrule
GASP & \bestnoreal{19.75$_{\pm2.33}$} & 79.67$_{\pm0.35}$ & \bestnoreal{70.63$_{\pm0.70}$} \\
\textsc{AZR} & 17.13$_{\phantom{\pm0.00}}$ & \bestnoreal{\textbf{83.54$_{\phantom{\pm0.00}}$}} & 69.58$_{\phantom{\pm0.00}}$ \\
\midrule
Real-data RL & 18.21$_{\pm1.41}$ & 78.66$_{\pm1.61}$ & 69.31$_{\pm0.95}$ \\
GASP + Real-data RL & \bestreal{\textbf{20.52$_{\pm3.60}$}} & \bestreal{80.49$_{\pm1.61}$} & \bestreal{\textbf{71.60$_{\pm1.00}$}} \\
\midrule
Qwen2.5-Coder-7B & 13.89$_{\phantom{\pm0.00}}$ & 79.88$_{\phantom{\pm0.00}}$ & 69.58$_{\phantom{\pm0.00}}$ \\
\bottomrule
\end{tabular}
\end{table}

\subsection{Ablation: Rejection Sampling in \ours}
\label{sec:abl_rejection_sampling}
\begin{figure*}[!h]
    \centering
    \includegraphics[width=0.7\linewidth]{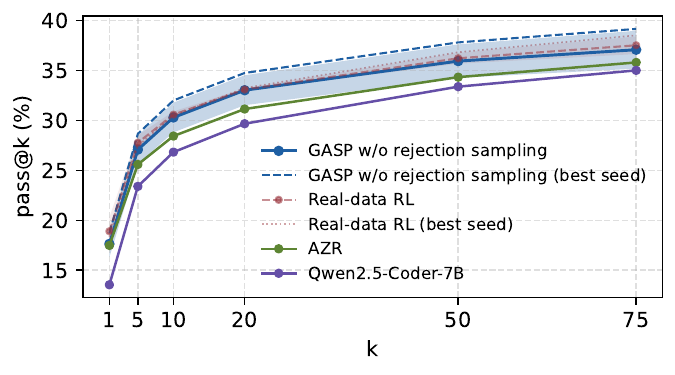}
    \caption{\textbf{Pass@k performance on the LCB eval benchmark (LCB\textsuperscript{v5}) for \ours without rejection sampling.} We compare \ours without rejection-sampling, Real-data RL, \textsc{AZR}, and \base. All RL-based results are repeated over three seeds. For each seed of our RL runs, we select the checkpoint with the best pass@20 within that run, and report mean $\pm$ std across seeds. For \textsc{AZR}, we evaluate the authors' publicly released checkpoint (trained from \base); the authors report selecting the best checkpoint over training from a single-seed run.}
    \label{fig:gasp_no_rejection_sampling}
\end{figure*}

\begin{figure*}[!h]
    \centering
    \includegraphics[width=0.7\linewidth]{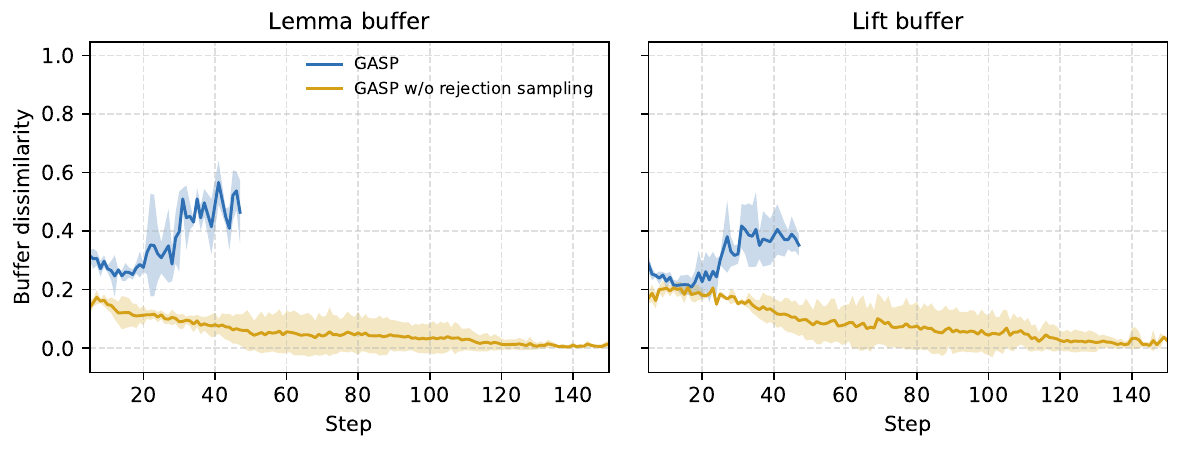}
    \caption{\textbf{Dissimilarity of proposed lemma and lift questions with respect to their corresponding buffers for \ours with and without rejection sampling.} We observe that \ours without rejection sampling suffers from diversity collapse as RL training progresses, with proposed questions becoming increasingly similar to those already in the buffer. Results shown for three seeds. (see \app{sec:buffer_dis} for more on the buffer dissimilarity computation.)}
    \label{fig:buffer_dis}
\end{figure*}

We showcase performance of \ours without rejection sampling in \fig{fig:gasp_no_rejection_sampling}. Compared to \ours with rejection sampling, we observe high variance among seeds, while still seeing improved performance over unguided self-play with \textsc{AZR}.
We attribute the high variance to the diversity collapse shown in \fig{fig:buffer_dis}: the timing of this collapse varies across runs, with seeds that maintain diversity longer achieving stronger benchmark performance.
Interestingly, \ours without rejection sampling solves more unique goalpost questions across seeds (14 vs.\ 11; see \tab{tab:hard-evals}), suggesting that concentrated training on similar problem types can benefit specific hard tasks even as general benchmark consistency suffers.

\subsection{Ablation: Removing i/o Difficulty Axis in \ours}
\label{sec:io-ablation}

\begin{figure*}[!h]
    \centering
    \includegraphics[width=0.7\linewidth]{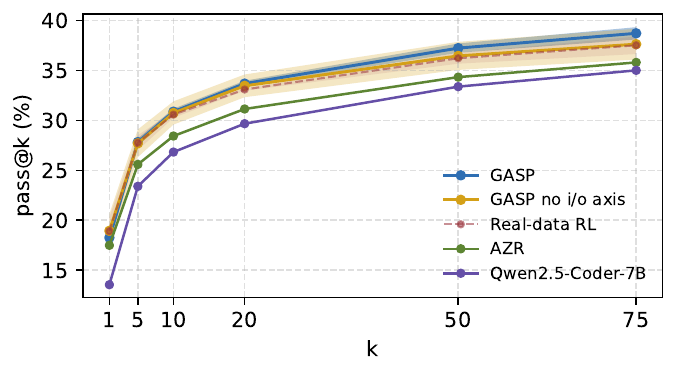}
    \caption{\textbf{Pass@k performance on the LCB eval benchmark (LCB\textsuperscript{v5}) for \ours without input-output (i/o) difficulty axis for the teacher proposals.} We compare \ours without the i/o difficulty axis (only relying on functional difficulty axis) with Real-data RL, \textsc{AZR}, and \base. All RL-based results are repeated over three seeds. For each seed of our RL runs, we select the checkpoint with the best pass@20 within that run, and report mean $\pm$ std across seeds. For \textsc{AZR}, we evaluate the authors' publicly released checkpoint (trained from \base); the authors report selecting the best checkpoint over training from a single-seed run.}
    \label{fig:gasp_no_io_axis}
\end{figure*}

We report the performance of \ours without the input-output (i/o) difficulty axis in \fig{fig:gasp_no_io_axis}.
Compared to \ours where we adjust difficulty along both the functional and input-output axes, we observe comparable performance, but higher variance among seeds.
We hypothesize that the i/o axis helps diversify the synthetic training distribution, resulting in more varied data and more consistent performance across seeds.

\subsection{Ablation: One-step Guidance in \ours}
\label{sec:one-step-ablation}
  We ablate the two-step lemma–lift curriculum by training with only one difficulty level: either medium or hard.

  \paragraph{One-step ablation rewards.}
    \begin{wrapfigure}[8]{r}{0.35\textwidth}%
  \vspace{-1\baselineskip}
    \includegraphics[width=1\linewidth]{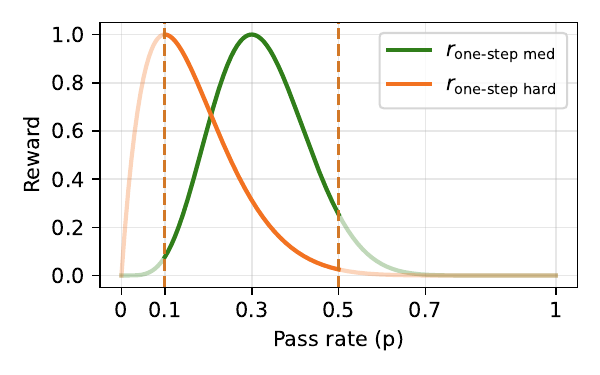}
  \caption{\textbf{Teacher rewards for \textcolor{lemmagreen}{one-step medium} and \textcolor{liftorange}{one-step hard} proposals.}
  }
  \label{fig:rewards_onestep}
  \vspace{-1.5\baselineskip}
\end{wrapfigure}
The one-step ablations replace the two-stage lemma--lift
curriculum with a single reward using the generalized form $\mathcal{R}(p;\, a, b)$ from \eqn{eq:generalized_reward}:
  \begin{align}
  r_{\text{one-step med}} &=
  \begin{cases}
  \mathcal{R}(p;\, 0.3,\, 5), & \text{if } 0.1 \leq p \leq 0.5 \\
  -0.5, & \text{otherwise,}
  \end{cases}
  \label{eq:onestep_med_reward} \\[6pt]
  r_{\text{one-step hard}} &=
  \begin{cases}
  \mathcal{R}(p;\, 0.1,\, 1), & \text{if } 0.1 \leq p \leq 0.5 \\
  -0.5, & \text{otherwise.}
  \end{cases}
  \label{eq:onestep_hard_reward}
  \end{align}

  The one-step medium reward peaks at $p = 0.3$, targeting intermediate difficulty, while the one-step hard reward peaks at $p = 0.1$, matching the lift reward. Both use the same reward band $[0.1, 0.5]$, as shown in \fig{fig:rewards_onestep}.

  \paragraph{Results} Both one-step variants underperform \ours on LCB (\tab{tab:lcb-joint-ablations}), with the gap most
  visible at pass@1. The one-step hard variant achieves slightly better benchmark performance than one-step medium, despite solving far fewer goalpost questions (2 vs.\ 11, \tab{tab:hard-evals}). This is consistent with the hypothesis that harder training questions benefit downstream evaluation, while easier lemma questions serve as stepping stones toward the goalposts. The two-step curriculum appears to combine both effects, though further investigation is needed to disentangle their individual contributions more precisely.

\begin{figure*}[!h]
    \centering
    \includegraphics[width=0.7\linewidth]{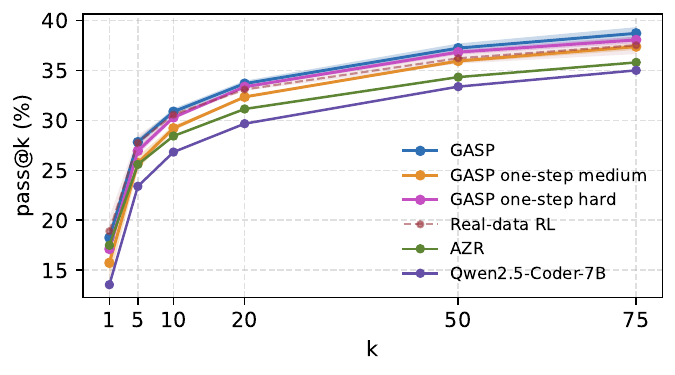}
    \caption{\textbf{Pass@k performance on the LCB eval benchmark
(LCB\textsuperscript{v5}) for \ours with one-step instead of the
two-step lemma and lift generation.} We compare \ours with a
one-step curriculum (intermediate difficulty: one-step medium,
hard: one-step hard), Real-data RL, \textsc{AZR}, and \base.
All RL-based results are repeated over three seeds. For each seed
of our RL runs, we select the checkpoint with the best pass@20
within that run, and report mean $\pm$ std across seeds. For
\textsc{AZR}, we evaluate the authors' publicly released checkpoint
(trained from \base); the authors report selecting the best
checkpoint over training from a single-seed run.}
    \label{fig:gasp_one_step}
\end{figure*}

\subsection{Solved Goalpost Questions} \label{app:goalpost}

\begin{figure*}[!h]
    \centering
    \includegraphics[width=0.6\linewidth]{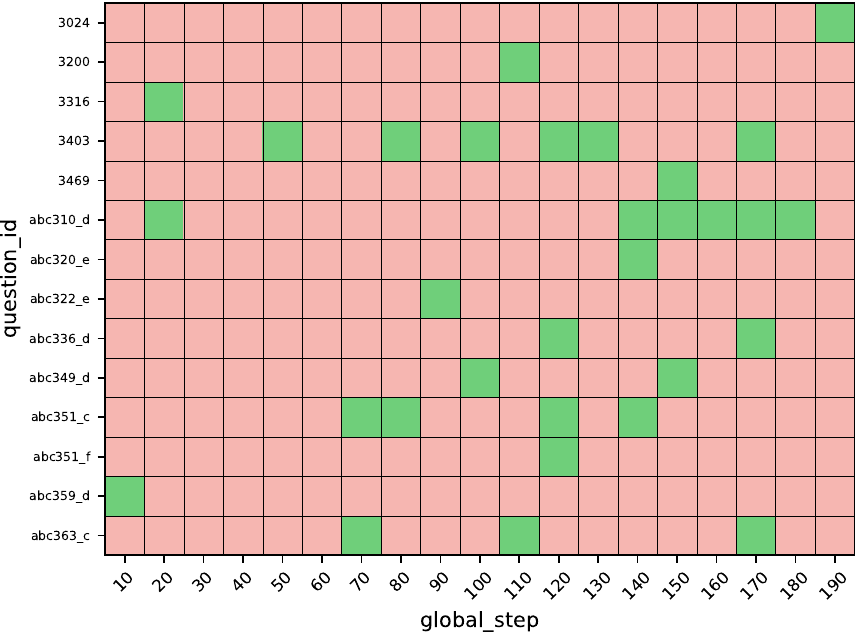}
    \caption{\textbf{Goalpost questions solved by \ours\ without rejection sampling.} Visualization over training checkpoints (x-axis: global step) and goalpost question IDs (y-axis). We include only goalpost questions that are solved at least once during training. A cell is marked as solved (green) if the corresponding question is solved at that checkpoint; otherwise it is unsolved (red). Results are aggregated across three seeds, where a cell is marked solved if \emph{any} seed solves the question at that checkpoint (union across seeds).}
\label{fig:gasp_no_rj_goalpost}
\end{figure*}

We report the number of unique goalpost questions in $\mathcal{H}$ solved across training checkpoints in \tab{tab:hard-evals}. A goalpost is counted as solved if pass@100 $> 0$ at any checkpoint within a run, counted at most once per seed.
Although \ours without rejection sampling solves 14 unique goalposts across seeds (5/8/5) versus 11 for \ours (6/3/3), it also trains for nearly 4$\times$ as many steps (190 vs.\ 48), giving it substantially more opportunities to encounter goalpost questions. We additionally provide the goalpost-solve plot for the \ours\ variant without rejection sampling in \fig{fig:gasp_no_rj_goalpost}.
As discussed in \app{sec:abl_rejection_sampling}, without rejection sampling, more and more similar lemma and lift questions are proposed by the teacher and are then subsequently solved by the student. We hypothesize that these similar questions might result in a more targeted effort toward the goalpost questions in some instances; however, this comes at the cost of high seed variance on downstream benchmarks.
When truncated to the same step range as \ours, \ours without rejection sampling solves only 3 unique goalposts compared to \ours's 11, suggesting that rejection sampling enables more sample-efficient discovery of hard solutions.

\begin{table}[t]
\centering
\caption{Unique goalpost questions solved across training checkpoints (out of 146 goalposts). A question is counted if it is solved at any checkpoint within a run (pass@100 $> 0$). Per-seed counts and union
across three seeds are reported. \textbf{Bold}: global best. $\dagger$Zero by construction (goalposts filtered to be unsolvable by these methods).}
\label{tab:hard-evals}
\begin{tabular}{clcc}
\toprule
& Method & Per-seed & Unique (union) \\
\midrule
 & GASP & 6 / 3 / 3 & 11 \\
\ldelim\{{4}{30pt}[\scriptsize\textit{Ablations}] & GASP no i/o axis & 2 / 2 / 2 & 4 \\
 & GASP w/o rejection sampling & 5 / 8 / 5 & \textbf{14} \\
 & GASP one-step medium & 6 / 4 / 4 & 11 \\
 & GASP one-step hard & 0 / 1 / 1 & 2 \\
 & \textsc{AZR} & --- & 0$^\dagger$ \\
\midrule
 & Real-data RL & --- & 0$^\dagger$ \\
 & GASP + Real-data RL & 9 / 2 / 3 & 10 \\
\midrule
\scriptsize\textit{Base} & Qwen2.5-Coder-7B & --- & 0$^\dagger$ \\
\bottomrule
\end{tabular}
\end{table}

\begin{figure}[t]
\centering

\begin{tikzpicture}[
  font=\small,
  box/.style={draw, rounded corners=2mm, align=left, inner sep=6pt, text width=0.92\linewidth},
  hdr/.style={font=\bfseries\footnotesize}
]

\node[box, draw=hardred, fill=hardfill, anchor=north] (hard2) {
  {\hdr\textcolor{hardred}{Goalpost (Hard)}}\hfill\texttt{hard\_id: abc364\_e}\quad\texttt{axis: io}\\[-2pt]
  \textcolor{hardred}{\rule{\linewidth}{0.3pt}}\\[4pt]
Takahashi has prepared N dishes for Snuke.
The dishes are numbered from 1 to N, and dish i has a sweetness of A\_i and a saltiness of B\_i.
Takahashi can arrange these dishes in any order he likes. Snuke will eat the dishes in the order they are arranged, but if at any point the total sweetness of the dishes he has eaten so far exceeds X or the total saltiness exceeds Y, he will not eat any further dishes.
Takahashi wants Snuke to eat as many dishes as possible.
Find the maximum number of dishes Snuke will eat if Takahashi arranges the dishes optimally.
};

\node[box, draw=lemmagreen, fill=lemmafill, below=3mm of hard2] (lemma2) {
  {\hdr\textcolor{lemmagreen}{Lemma}}\hfill\texttt{axis: io}\\[-2pt]
  \textcolor{lemmagreen}{\rule{\linewidth}{0.3pt}}\\[4pt]
Snuke has a limit for either the sweetness or saltiness, and he wants to know the maximum number of options he can choose from without exceeding the limit. Given a limit X and a list of options with sweetness and saltiness, determine the maximum number of options he can choose.
};

\node[box, draw=liftorange, fill=liftfill, below=3mm of lemma2] (lift2) {
  {\hdr\textcolor{liftorange}{Lift}}\hfill\texttt{axis: io}\\[-2pt]
  \textcolor{liftorange}{\rule{\linewidth}{0.3pt}}\\[4pt]
Snuke now has two limits, one for the sweetness of food and another for the saltiness. He wants to know the maximum number of options he can choose without either of these limits being exceeded. Given two limits for sweetness and saltiness, and a list of options with their sweetness and saltiness, determine the maximum number of options he can choose without exceeding either limit.
};

\draw[->, thick, hardred!70] (hard2.south) -- (lemma2.north);
\draw[->, thick, lemmagreen!70] (lemma2.south) -- (lift2.north);

\end{tikzpicture}

\vspace{1cm}

\begin{tikzpicture}[
  font=\small,
  box/.style={draw, rounded corners=2mm, align=left, inner sep=6pt, text width=0.92\linewidth},
  hdr/.style={font=\bfseries\footnotesize}
]

\node[box, draw=hardred, fill=hardfill, anchor=north] (hard4) {
  {\hdr\textcolor{hardred}{Goalpost (Hard)}}\hfill\texttt{hard\_id: abc358\_e}\quad\texttt{axis: io}\\[-2pt]
  \textcolor{hardred}{\rule{\linewidth}{0.3pt}}\\[4pt]
AtCoder Land sells tiles with English letters written on them. Takahashi is thinking of making a nameplate by arranging these tiles in a row. Find the number, modulo 998244353, of strings consisting of uppercase English letters with a length between 1 and K, inclusive, that satisfy the following conditions: - For every integer i satisfying $1 < i < 26$, the following holds: - Let a\_i be the i-th uppercase English letter in lexicographical order. - The number of occurrences of a\_i in the string is between 0 and C\_i, inclusive.
};

\node[box, draw=lemmagreen, fill=lemmafill, below=3mm of hard4] (lemma4) {
  {\hdr\textcolor{lemmagreen}{Lemma}}\hfill\texttt{axis: io}\\[-2pt]
  \textcolor{lemmagreen}{\rule{\linewidth}{0.3pt}}\\[4pt]
Design a Python function that takes two arguments: a string of characters and an integer representing the length of the strings to be generated. The function should return the number of unique strings of the given length that can be formed using the characters in the input string. Note that the order of characters matters, and each character can be used any number of times.
};

\node[box, draw=liftorange, fill=liftfill, below=3mm of lemma4] (lift4) {
  {\hdr\textcolor{liftorange}{Lift}}\hfill\texttt{axis: io}\\[-2pt]
  \textcolor{liftorange}{\rule{\linewidth}{0.3pt}}\\[4pt]
Design a Python function that takes two arguments: a string of characters and an integer representing the length of the strings to be generated. The function should return the number of unique palindromic strings of the given length that can be formed using the characters in the input string. Note that each character can be used any number of times, and the order of characters matters.
};

\draw[->, thick, hardred!70] (hard4.south) -- (lemma4.north);
\draw[->, thick, lemmagreen!70] (lemma4.south) -- (lift4.north);

\end{tikzpicture}

\caption{\textbf{Goalpost$\rightarrow$lemma$\rightarrow$lift examples from \ours run (with rejection sampling).} Input--output-axis (io) selected for difficulty adjustment on hard questions with id \texttt{abc364\_e} (top) and \texttt{abc358\_e} (bottom). Shared motifs: for \texttt{abc364\_e} maximizing count under cumulative resource constraints (1D$\rightarrow$2D feasibility); for \texttt{abc358\_e} counting strings under per-letter usage caps (mod / DP).}
\label{fig:lemma-lift-more}
\end{figure}
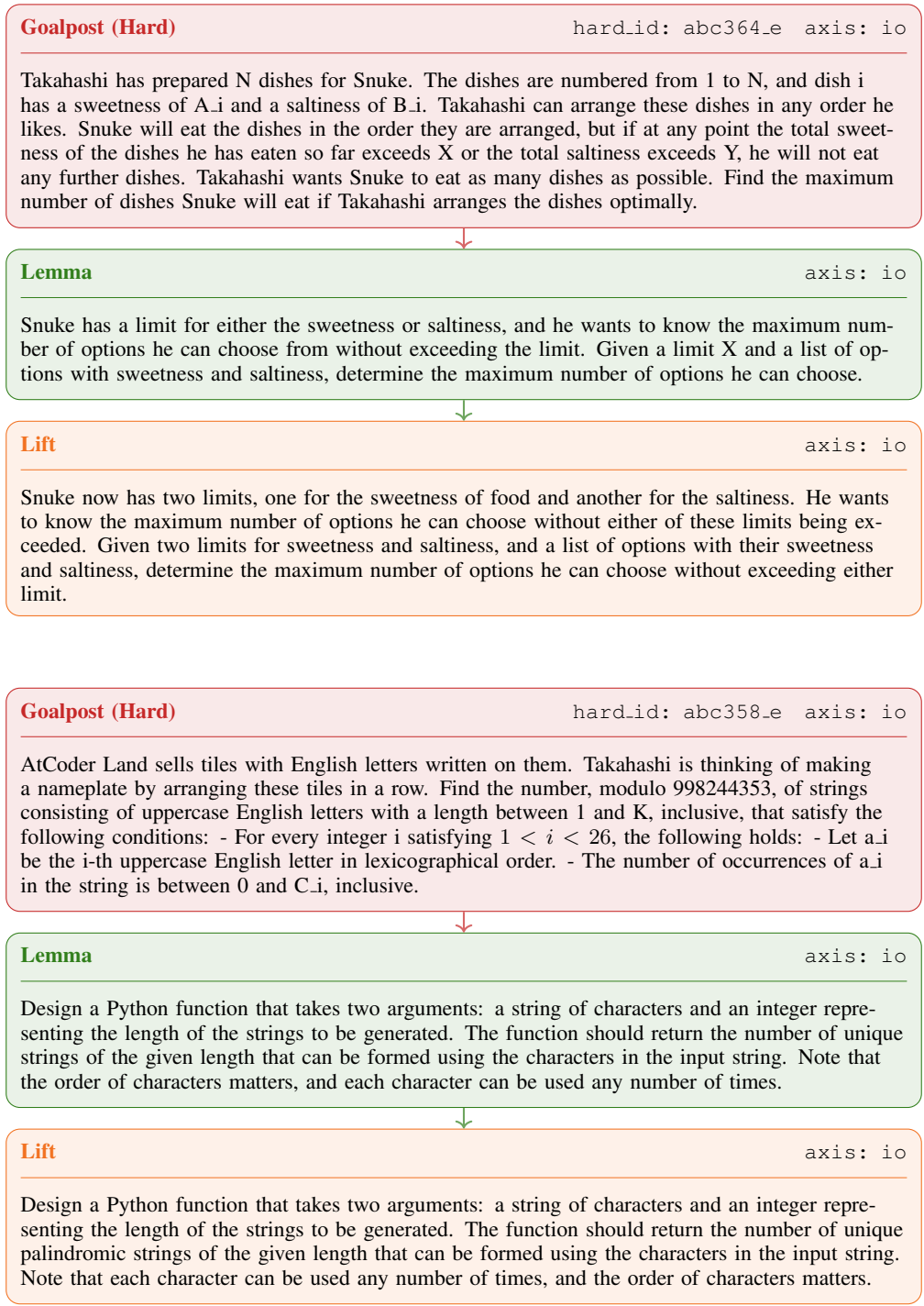

\begin{figure}[t]
\centering

\begin{tikzpicture}[
  font=\small,
  box/.style={draw, rounded corners=2mm, align=left, inner sep=6pt, text width=0.92\linewidth},
  hdr/.style={font=\bfseries\footnotesize}
]

\node[box, draw=hardred, fill=hardfill, anchor=north] (hard3) {
  {\hdr\textcolor{hardred}{Goalpost (Hard)}}\hfill\texttt{hard\_id: 3414}\quad\texttt{axis: f}\\[-2pt]
  \textcolor{hardred}{\rule{\linewidth}{0.3pt}}\\[4pt]
You are given a non-negative integer k. There exists a staircase with an infinite number of stairs, with the lowest stair numbered 0. Alice has an integer jump, with an initial value of 0. She starts on stair 1 and wants to reach stair k using any number of operations. If she is on stair i, in one operation she can: Go down to stair i - 1. This operation cannot be used consecutively or on stair 0. Go up to stair i + $2^{jump}$. And then, jump becomes jump + 1. Return the total number of ways Alice can reach stair k.
};

\node[box, draw=lemmagreen, fill=lemmafill, below=3mm of hard3] (lemma3) {
  {\hdr\textcolor{lemmagreen}{Lemma}}\hfill\texttt{axis: f}\\[-2pt]
  \textcolor{lemmagreen}{\rule{\linewidth}{0.3pt}}\\[4pt]
You are given a non-negative integer k. You start at stair 1 and want to reach stair k using any number of operations. In one operation, you can either: 1. Move down to the previous stair (i - 1). 2. Move up to the next two positions (i + 1). Your task is to write a function that returns the total number of ways you can reach stair k starting from stair 1.
};

\node[box, draw=liftorange, fill=liftfill, below=3mm of lemma3] (lift3) {
  {\hdr\textcolor{liftorange}{Lift}}\hfill\texttt{axis: f}\\[-2pt]
  \textcolor{liftorange}{\rule{\linewidth}{0.3pt}}\\[4pt]
You are given a non-negative integer k and a positive integer moves\_limit. You start at stair 0 and want to reach stair k using any number of operations, but you are limited by the moves\_limit. In one operation, you can either: 1. Move down to the previous stair (i - 1). 2. Move up to the next two positions (i + 1). Your task is to write a function that returns the total number of ways you can reach stair k starting from stair 0, given the moves\_limit.
};

\draw[->, thick, hardred!70] (hard3.south) -- (lemma3.north);
\draw[->, thick, lemmagreen!70] (lemma3.south) -- (lift3.north);

\end{tikzpicture}

\caption{\textbf{Goalpost$\rightarrow$lemma$\rightarrow$lift examples from \ours run (with rejection sampling).} Function-axis (f) selected for difficulty adjustment on hard question with id \texttt{3414}. Shared motif: counting paths / number of ways in a stateful process with constrained moves; DP on position + auxiliary state.}
\label{fig:lemma-lift-more-f}
\end{figure}
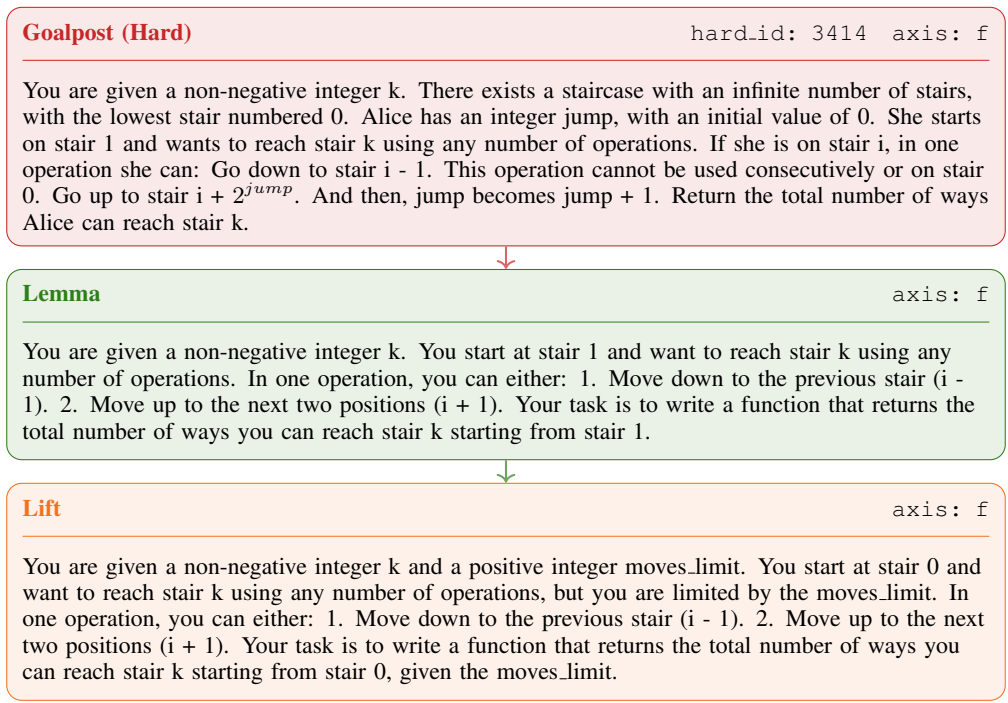

 \section{Hyperparameter Settings}
  \paragraph{Shared settings.}

  All experiments use a batch size of 64, a learning rate of $10^{-6}$, and a temperature of 1.0 for training and 0.6 for evaluation. All training is performed on the \texttt{LiveCodeBench} training split (pre-2024.08),
   with evaluations on the LCB\textsuperscript{v5} split as well as HumanEval$^+$, and MBPP$^+$. All experiments use the vLLM~\citep{kwon2023efficient} and verl~\citep{sheng2024hybridflow} frameworks for distributed training and evaluation. Following
  AZR, we use 4$\times$ H100 GPUs for training and 1$\times$ H100 for evaluation, with code execution managed through the same process as AZR.

  \paragraph{RLVR.}
  RLVR training uses GRPO for 500 steps with 3 seeds. For each prompt, $n{=}8$ responses are sampled for advantage calculation. We observe saturation of pass@$k$ within 200 steps and a loss of generalization beyond that.

  \paragraph{GASP.}
  GASP experiments run for approximately 50 global iteration steps, with evaluations every 2 steps. Each global step involves lemma generation with rejection sampling, lift generation with rejection sampling, and a solver update. Each
  rejection sampling step can contain multiple generation steps until conditions are met or \texttt{max\_attempts} is reached, with one model update per rejection sampling step. As described in \alg{alg:gasp-training}, there can be multiple
  lemma generation steps within one global iteration until the minimum count of total generated programs per step is met. GASP models are trained using Task-Relative REINFORCE++, following \textsc{AZR}~\citep{zhao2025absolute}. For comparison, the authors of AZR report training for at
  least approximately 350 steps.

  Ablations without rejection sampling run for approximately 200 global iteration steps, with evaluations every 10 steps. In both cases, evaluation performance had not yet saturated, suggesting room for further improvement.

  \paragraph{Solver.}
  We set $N{=}10$ for estimating the solver's success rate, based only on induction tasks (not abduction or deduction). For each task, we obtain 5 example input-output pairs: 2 are shown as public examples and 3 are held out for private
  testing to avoid overfitting.

\end{document}

%% file: math_commands.tex
\usepackage{amsmath,amsfonts,bm}

\def\eqref#1{equation~\ref{#1}}

\def\1{\bm{1}}

\DeclareMathAlphabet{\mathsfit}{\encodingdefault}{\sfdefault}{m}{sl}
\SetMathAlphabet{\mathsfit}{bold}{\encodingdefault}{\sfdefault}{bx}{n}



%% file: al_header.tex
\usepackage{booktabs}       %

\usepackage{graphics}
\usepackage{mathtools}

\usepackage{enumitem}

\usepackage{amsfonts}       %
\usepackage{amsmath}       %
\usepackage{amssymb}
\usepackage{amsthm}
\usepackage{pifont}%
\newcommand{\cmark}{\ding{51}}%
\newcommand{\xmark}{\ding{55}}%
\usepackage{diagbox}
\usepackage{multirow}

\makeatletter
\newcommand\notsotiny{\@setfontsize\notsotiny{6.31415}{7.1828}}
\makeatother

\newcommand{\Fig}[1]{Figure~\ref{#1}}  %
\newcommand{\fig}[1]{Fig.~\ref{#1}}    %

\newcommand{\alg}[1]{Alg.~\ref{#1}}
\newcommand{\tab}[1]{Table~\ref{#1}}

\newcommand{\eqn}[1]{Eq.~\ref{#1}} %
\renewcommand{\sec}[1]{Sec.~\ref{#1}} %
\newcommand{\supp}[1]{Suppl.~\ref{#1}}
\newcommand{\app}[1]{Appendix.~\ref{#1}}

\usepackage{xspace}
\makeatletter
\DeclareRobustCommand\onedot{\futurelet\@let@token\@onedot}
\def\@onedot{\ifx\@let@token.\else.\null\fi\xspace}
\def\eg{e.g\onedot}
\def\ie{i.e\onedot}

\makeatother

\newcommand{\Real}{\ensuremath{\mathbb R}}        %

\let\originalleft\left
\let\originalright\right
\renewcommand{\left}{\mathopen{}\mathclose\bgroup\originalleft}
\renewcommand{\right}{\aftergroup\egroup\originalright}

\newcommand{\g}[1]{%
  \ifthenelse{\equal{#1}{(}}
  {\left( }%
    { \ifthenelse{\equal{#1}{)}}
      { \right)}%
    { \ifthenelse{\equal{#1}{[}}
      {\left[}%
        { \ifthenelse{\equal{#1}{]}}
          { \right]}%
        {#1}}
    }
  }
}

\definecolor{ourblue}{rgb}{0.368,0.507,0.71}
\definecolor{ourorange}{rgb}{0.881,0.611,0.142}
\definecolor{ourgreen}{rgb}{0.56,0.692,0.195}
\definecolor{ourred}{rgb}{0.923,0.386,0.209}
\definecolor{ourviolet}{rgb}{0.528,0.471,0.701}
\definecolor{ourbrown}{rgb}{0.772,0.432,0.102}
\definecolor{ourlightblue}{rgb}{0.364,0.619,0.782}
\definecolor{ourdarkgreen}{rgb}{0.572,0.586,0.}
\definecolor{ourdarkblue}{RGB}{28,99,148}
\definecolor{ourdarkred}{RGB}{169,53,17}

\usepackage{etoolbox}
\makeatletter
\patchcmd{\NAT@test}{\else \NAT@nm}{\else \NAT@nmfmt{\NAT@nm}}{}{}
\DeclareRobustCommand\citeposs
  {\begingroup
   \let\NAT@nmfmt\NAT@posfmt%
   \NAT@swafalse\let\NAT@ctype\z@\NAT@partrue
   \@ifstar{\NAT@fulltrue\NAT@citetp}{\NAT@fullfalse\NAT@citetp}}

\let\NAT@orig@nmfmt\NAT@nmfmt
\def\NAT@posfmt#1{\NAT@orig@nmfmt{#1's}}
\makeatother

\usepackage{xr-hyper}

\makeatletter
\newcommand*{\addFileDependency}[1]{%
  \typeout{(#1)}
  \@addtofilelist{#1}
  \IfFileExists{#1}{}{\typeout{No file #1.}}
}
\makeatother

%% file: color_header.tex
\definecolor{springreen}{RGB}{225, 247, 208}
\definecolor{outlinegreen}{RGB}{145, 170, 126}
\definecolor{boxyellow}{RGB}{245, 238, 156}
\definecolor{darkyellow}{RGB}{145, 138, 56}
\definecolor{lightpink}{RGB}{249, 203, 203}
\definecolor{gateclosed}{RGB}{107, 107, 107}
\definecolor{greyish}{RGB}{100, 100, 100}
\definecolor{pastelpurple}{RGB}{199, 206, 234}
\definecolor{darkpurple}{RGB}{99, 106, 134}
\definecolor{pastelorange}{RGB}{255, 218, 193}
\definecolor{darkorange}{RGB}{155, 118, 93}
\definecolor{pastelblue}{RGB}{192, 228, 241}
\definecolor{darkblue}{RGB}{92, 128, 141}
\definecolor{pastelred}{RGB}{255, 154, 162}
\definecolor{darkred}{RGB}{155, 54, 62} %
\definecolor{lightgrey}{RGB}{230, 230, 230}
\definecolor{wmblue}{RGB}{0, 146, 179}
\definecolor{darkwmblue}{RGB}{38, 112, 130} %
\definecolor{darkbrown}{RGB}{132, 119, 113} %
\definecolor{darkwmgreen}{RGB}{93, 133, 51} %
\definecolor{gptgreen}{RGB}{12, 163, 128}
\definecolor{newpurple}{RGB}{101, 79, 167} %
\definecolor{pastelpurple}{RGB}{171,159,209}
\definecolor{lightpurple}{RGB}{169, 176, 204}
\definecolor{newlightpurple}{RGB}{169, 181, 214}

\definecolor{midgrey}{RGB}{200, 200, 200}%
\definecolor{darkgreen}{rgb}{0.568627450980392,0.666666666666667,0.494117647058824}

\definecolor{robodeskorange}{RGB}{253, 161, 85}
\definecolor{greyish}{RGB}{100, 100, 100}

\definecolor{pink}{RGB}{255, 51, 155}